%% file: main.tex
\documentclass{article} 
\usepackage{iclr2026_conference,times}

\input{math_commands.tex}

\usepackage{hyperref}
\usepackage{url}
\usepackage{graphicx}
\usepackage{placeins}

\usepackage[most]{tcolorbox}
\usepackage{wrapfig}
\usepackage[table]{xcolor}

\usepackage{xcolor}
\usepackage{seqsplit}
\usepackage{amssymb}
\usepackage{pifont}
\usepackage{booktabs}
\usepackage{listings}
\usepackage{multirow}

\definecolor{boxbg}{RGB}{240, 240, 250}
\definecolor{takeawaycolor}{RGB}{70, 70, 180} 

\newtcolorbox{takeaway}{%
  enhanced,
  rounded corners,
  breakable,
  colback=boxbg,
  colframe=boxbg,
  boxrule=0pt,
  left=10pt,
  right=10pt,
  top=5pt,
  bottom=5pt,
  toptitle=1pt,
  bottomtitle=1pt,
  arc= 12pt,
  before={\vspace{0pt}},
  after={\vspace{0pt}}
}

\newtcolorbox{takeaway2}{ 
    boxrule = 1.5pt,
    rounded corners,
    arc = 2pt   
}

\title{WebCoach: Self-Evolving Web Agents with Cross-Session Memory Guidance}

\author{Genglin Liu \thanks{Work done while interning at Amazon. Code available at \url{https://github.com/genglinliu/WebCoach}}\\
University of California, Los Angeles\\
\And
Shijie Geng \\
Amazon\\
\And
Sha Li \\
Amazon\\
\AND
Hejie Cui \\
Amazon\\
\And
Sarah Zhang \\
Amazon\\
\And
Xin Liu \\
Amazon\\
\And
Tianyi Liu \\
Amazon
}

\iclrfinalcopy 
\begin{document}

\maketitle

\input{sections/0_abstract}

\input{sections/1_intro}

\input{sections/2_framework}

\input{sections/3_experiments}

\input{sections/4_results}

\input{sections/5_related_work}

\input{sections/6_conclusion}

\bibliography{iclr2026_conference}
\bibliographystyle{iclr2026_conference}

\appendix

\input{sections/7_appendix}
\end{document}

%% file: math_commands.tex
\usepackage{amsmath,amsfonts,bm}

\def\eqref#1{equation~\ref{#1}}

\def\1{\bm{1}}

\DeclareMathAlphabet{\mathsfit}{\encodingdefault}{\sfdefault}{m}{sl}
\SetMathAlphabet{\mathsfit}{bold}{\encodingdefault}{\sfdefault}{bx}{n}



%% file: sections/0_abstract.tex
\begin{abstract}
    Multimodal LLM-powered agents have recently demonstrated impressive capabilities in web navigation, enabling agents to complete complex browsing tasks across diverse domains. However, current agents struggle with repetitive errors and lack the ability to learn from past experiences across sessions, limiting their long-term robustness and sample efficiency.~We introduce \textbf{WebCoach}, a model-agnostic self-evolving framework that equips web browsing agents with persistent cross-session memory, enabling improved long-term planning, reflection, and continual learning without retraining.
    WebCoach consists of three key components: (1) a \textit{WebCondenser}, which standardizes raw navigation logs into concise summaries;  (2) an \textit{External Memory Store}, which organizes complete trajectories as episodic experiences; and (3) a \textit{Coach}, which retrieves relevant experiences based on similarity and recency, and decides whether to inject task-specific advice into the agent via runtime hooks. This design empowers web agents to access long-term memory beyond their native context window, improving robustness in complex browsing tasks. Moreover, WebCoach achieves self-evolution by continuously curating episodic memory from new navigation trajectories, enabling agents to improve over time without retraining.
    Evaluations on the WebVoyager benchmark demonstrate that WebCoach consistently improves the performance of browser-use agents across three different LLM backbones. With a 38B model, it increases task success rates from 47\% to 61\% while reducing or maintaining the average number of steps. Notably, smaller base models with WebCoach achieve performance comparable to the same web agent using GPT-4o.
\end{abstract}

\begin{figure}[!ht]
    \centering
    \includegraphics[width=\linewidth]{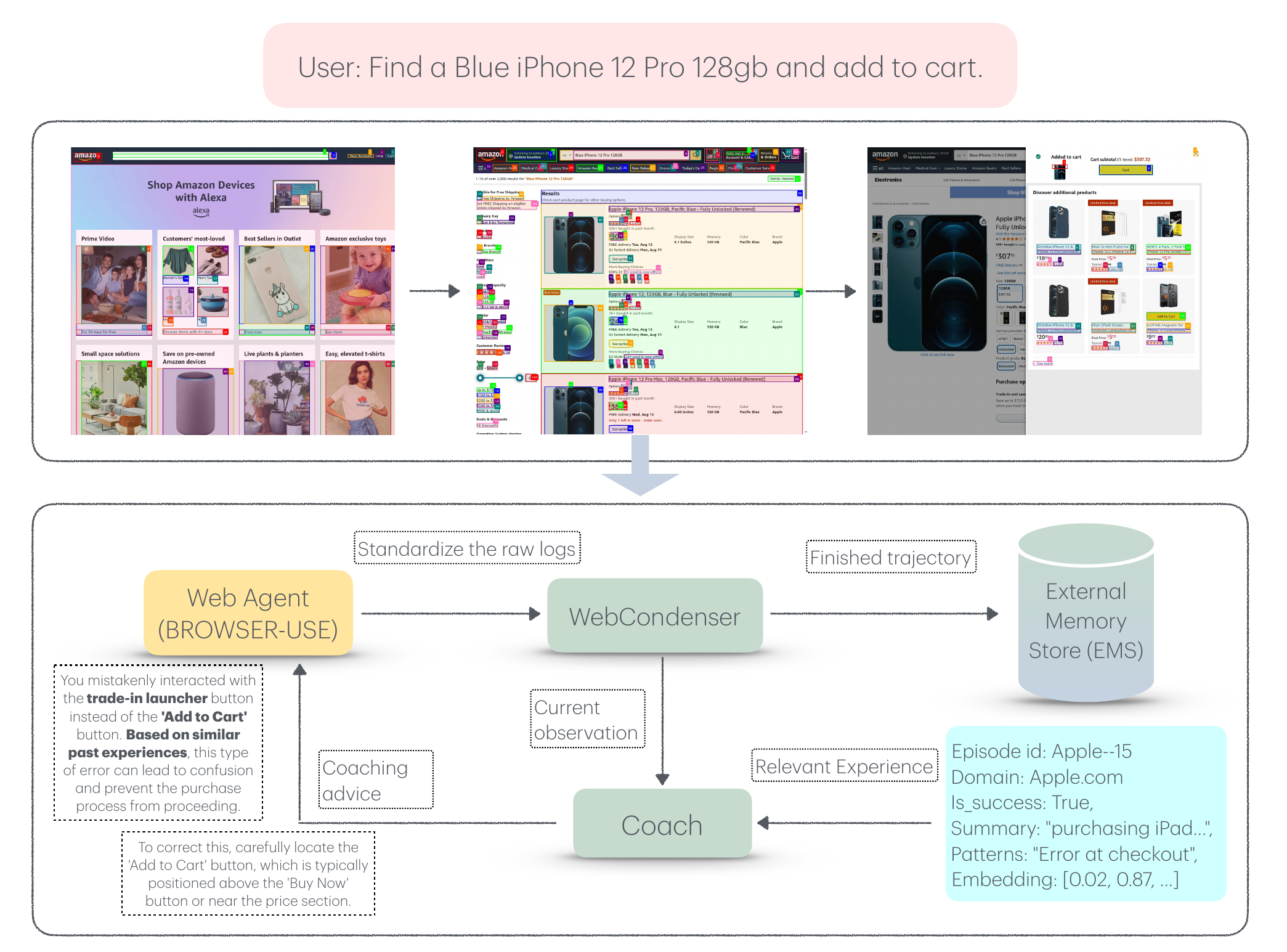}
    \vspace{-4pt}
    \caption{
    \textbf{Overview of the WebCoach framework.}
    WebCoach augments web-browsing agents with persistent, cross-session memory through an \textit{External Memory Store (EMS)} and a retrieval-augmented coaching mechanism. The \textit{Condenser} converts raw navigation histories into standardized summaries stored in \textit{EMS}, from which the \textit{Coach} retrieves relevant prior experiences to provide task-specific guidance to the main web agent. This design enables long-term planning, reflection, and continual improvement across browsing sessions.}
    \label{fig:webcoach_framework}
\end{figure}

%% file: sections/1_intro.tex
\section{Introduction}

Large language models (LLMs) have recently shown impressive skills in web and GUI navigation, enabling agents to fill forms, book flights, or compare shops across complex interfaces \citep{nakano2021webgpt, wei2025webagent, wu2025guiactor, zhang2025agentcpm, qin2025ui}.  
This rapid progress spans both desktop-style pages and mobile apps, with techniques ranging from coordinate-free visual grounding to reinforcement fine-tuning and multimodal tutorial mining \citep{luo2025gui, zhang2025tongui}.


Despite these advances, web agents still waste many steps: they revisit the same links, stall at login gates, or trigger CAPTCHAs across sessions \citep{li2025websailor, lyu2025deepshop, huang2025scaletrack}.  
Recent work introduces back-tracking or progress rewards to mitigate single-episode errors \citep{wu2025backtrackagent, zhang2025progrm}, but agents rarely \emph{remember} mistakes or successes beyond the current task. In short, most systems navigate without long-term memory -- an ability humans rely on to internalize past experiences and anticipate pitfalls.

The absence of memory fundamentally limits sample efficiency and robustness.  Even lightweight history compression or paging enhances web automation accuracy \citep{zhu2025turbocharging, kang2025memory}, while episodic reflection and contextual replay are known to improve adaptation in other domains \citep{shinn2023reflexion, liu2025contextual}.  However, these ideas have not been fully integrated into mainstream web-navigation pipelines.

We introduce \textbf{WebCoach}, a lightweight, model-agnostic framework that layers memory-aware guidance onto any existing web agent. WebCoach follows a simple premise: agents should learn from their own trajectories—successes, failures, and edge cases—\emph{without} retraining the base policy. It realizes this through three modules:  
(1) a \emph{WebCondenser} that converts raw interaction traces into compact semantic summaries;  
(2) an \emph{External Memory Store} that indexes and retrieves relevant episodes; and  
(3) a \emph{Coach} trainable LLM that decides when and how to intervene mid-episode. This architecture is framework-independent, as it wraps agents such as \textsc{browser-use} via simple trajectory hooks. Memories can be written online, shared across agents, and bootstrapped from curated traces, which avoids the lengthy reinforcement cycles demanded by prior improvements \citep{yin2025unlocking, xie2025gui}. As a result, even a small seed corpus yields useful advice from the first run.

On the WebVoyager benchmark \citep{he2024webvoyager}, which spans 643 live browsing tasks across 15 web domains, WebCoach consistently improves long-horizon success rates and efficiency across diverse base models.  
When paired with the \textsc{Skywork-38B} agent, WebCoach raises the overall success rate from \textbf{47\% to 61\%} -- a \textbf{14-point gain} -- while maintaining or reducing the average number of steps per task.  
Other models, such as \textsc{Qwen-VL-32B}, exhibit similar gains, increasing their success rate from 49\% to 57\% and thereby achieving performance comparable to GPT-4o with open-source backbones.  
Notably, dynamic self-experience memory (where agents iteratively expand their own memory store) outperforms externally seeded memories, highlighting that agents learn most effectively from their own trajectories rather than borrowed ones.  

In summary, this work advocates for a new, memory-centered paradigm in web agent design -- one where guidance is not hardcoded, but derived from experience. WebCoach is a step towards that goal, offering a practical and extensible architecture for building agents that truly remember and learn from their interactions.

%% file: sections/2_framework.tex
\section{WebCoach: A Model-Agnostic Framework for Memory-Augmented Web Navigation}
\label{sec:webcoach}

WebCoach is a lightweight, plug-and-play layer that augments any existing web-navigation agent (the \textit{actor}) with memory-aware guidance. It operates through three decoupled components: \textbf{WebCondenser}, an \textbf{External Memory Store (EMS)}, and the \textbf{Coach}. These components communicate via function calls and therefore require \emph{no} modifications to the actor's internal agentic workflow, enabling model-agnostic improvements on web browsing tasks.

\subsection{WebCondenser}
\label{ssec:condenser}

The WebCondenser is the first stage in the WebCoach framework and serves as the bridge between raw agent trajectories and memory-aware reasoning. Its role is deliberately narrow but essential: to convert low-level environment traces into semantically meaningful summaries. After every environment step, the actor agent logs a JSON file describing its current trajectory. This includes its observations, actions, and intermediate rewards (not necessarily a numerical score but self-evaluations on the status of the task completion). The WebCondenser parses these raw traces and condenses them into a structured schema that includes a concise natural language summary, a dense embedding, and key metadata such as whether the task was successfully completed. More details of the WebCondenser's operations and output formats are provided in Appendix~\ref{app:visual-example}.

\paragraph{Input.}
After every environment step, the actor writes a structured log describing the current partial trajectory $T_{1:t} = \bigl\{(o_i,a_i,r_i)\bigr\}_{i=1}^{t}$ (observation, action, reward).
WebCondenser is agnostic to the exact schema, allowing compatibility with frameworks like \textsc{browser-use}, Nova-Act, or any other framework.

\paragraph{Processing.}
A small LLM ($\le$ 8B) converts the raw trace into a fixed schema:

\begin{itemize}
    \item \texttt{summary\_text}: 3-5 sentences capturing the high-level outcome so far.
    \item \texttt{embedding}: 1536-d OpenAI embedding vector of the summary. This block could be replaced by other high quality embedding models as long as we keep the consistency across items in the EMS. 
    \item \texttt{final\_success}: \texttt{true}/\texttt{false}/\texttt{null} (task still running).
    \item \texttt{fail\_modes} or \texttt{success\_workflows}: Evidence with key steps for error analysis. If a complete trajectory successfully executed the user query and satisfied all the specified conditions, then the WebCondenser should highlight the workflows that led to this success; otherwise, it should summarize the errors or challenges that led to an unsuccessful attempt. 
\end{itemize}

The model also detects whether the trace is \textit{partial} or \textit{complete}. This process ensures fast runtime performance and compatibility across different frameworks. Importantly, the WebCondenser is not designed to perform any reasoning or intervention itself. Instead, it acts as a lightweight, schema-normalizing filter that prepares trajectory data for downstream modules. 

\paragraph{Routing.}

A crucial design decision is the distinction between partial and completed trajectories. While partial episodes (i.e., tasks still in progress) are streamed to the Coach for real-time decision-making, they are not stored. Only completed episodes that reach a natural stopping point are persisted to memory. This routing ensures the memory is populated with finalized examples that reflect either success or failure, preventing the accumulation of noise or transient decision states.

If the current step does not terminate the task, the record is streamed to the Coach for immediate use while \emph{skipping} storage.  Only once the task ends, the Condenser flags the trace as complete and persists it in the EMS; this prevents polluting memory with half-finished noise yet still gives the Coach a live context.

\subsection{External Memory Store (EMS)}
\label{ssec:ems}

The External Memory Store (EMS) is the long-term memory backbone of the WebCoach framework. It accumulates all completed browsing episodes, storing a semantic embedding of each agent \textit{experience} alongside the natural language summaries and metadata generated by the \textbf{WebCondenser}. This memory structure allows new browsing tasks to benefit from past experience, enabling efficient retrieval of relevant trajectories that reflect similar goals, domains, or interaction patterns.

\paragraph{Data schema.}
Each record stores
\(\bigl\langle\!\texttt{embedding}, \texttt{summary\_text}, \texttt{meta}\bigr\rangle\),
where \texttt{meta} information contains high-level identifications including
{\small\verb|episode_id|}, domain\,/\,URL root, user goal, model name, total steps, and timestamp.

\paragraph{Retrieval.}
The EMS is implemented as a vector database using FAISS with HNSW indexing \cite{douze2025faiss}, which supports efficient top-$K$ similarity search even as the memory grows to millions of episodes. Once candidate memories are retrieved, they are ranked by their similarity to the current context embedding, which is computed via a normalized dot product between the embedding vectors. HNSW builds a multi-layer navigable small-world graph that enables logarithmic-time approximate nearest-neighbor search with high recall \citep{malkov2018efficient}. In our implementation, we use FAISS with the HNSW-128.

\begin{wrapfigure}[18]{r}{0.45\textwidth}
\centering
  \includegraphics[width=0.45\textwidth]{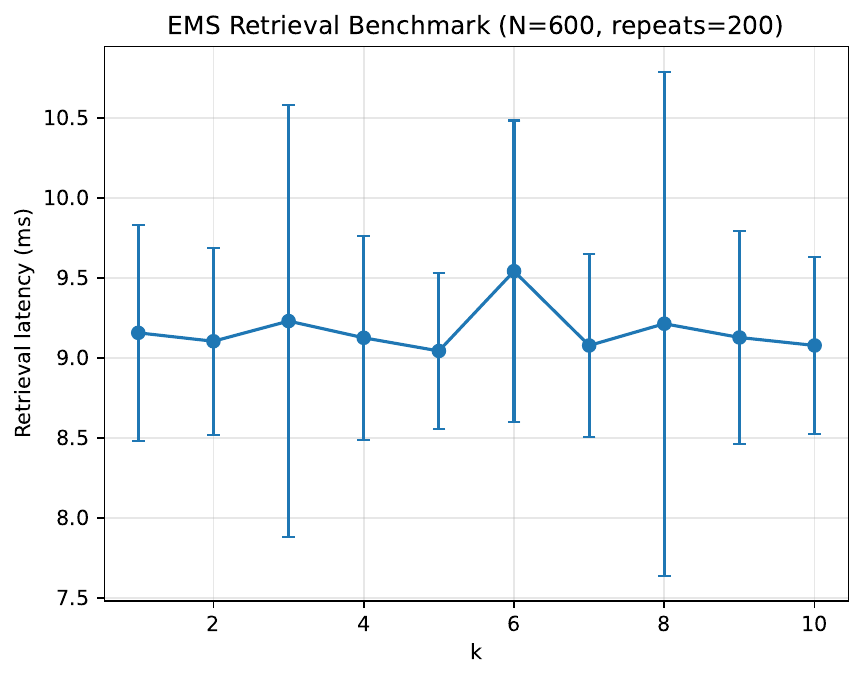}
  \caption{Retrieval speed at k for the EMS with 600 trajectories. Repeat each 200 times to measure the consistency. Most runs end up averaging between 9.0 and 9.5 ms for k ranging from 1 to 10.} 
  \label{fig:retrieval_speed}
\end{wrapfigure}

Given the Condenser embedding $\mathbf{e}_t$ for the current partial trace, EMS returns the top-$K$ past experiences using
\[
    \operatorname{score}(\mathbf{e}_t, \mathbf{e}_i)
    \;=\;
    \frac{\mathbf{e}_t^\top \mathbf{e}_i}{\|\mathbf{e}_t\|_2 \, \|\mathbf{e}_i\|_2} \, .
\]

\paragraph{Cold start.}
A key feature of the EMS is its generality: it is agnostic to the model, dataset, or domain an episode originated from. This universal design allows it to serve as a cross-actor, cross-task knowledge repository and enables bootstrapping. During a cold start phase, the EMS can be seeded with high-quality episodes from previously trained web agents, ensuring the Coach has relevant experience to draw upon from the very first online episode.

\subsection{Coach}
\label{ssec:coach}

The Coach is the reasoning engine of WebCoach. Unlike the WebCondenser and EMS, which deal with summarization and storage respectively, the Coach operates at runtime to enhance decision-making for the actor agent. At each step, the Coach receives a real-time summary of the actor’s current partial trajectory along with a set of retrieved memories from the EMS. It then decides whether to intervene by providing the actor with additional guidance. It is implemented by a 8B LLM that decides whether to inject additional context into the actor’s next prompt.

\paragraph{Inputs.} The Coach leverages both the content and the outcome labels (success/failure) of past experiences to issue advice that is grounded in evidence. For example, it might say: “Avoid clicking ‘Next’—previous agents got stuck in a loop here,” or “Try clicking the first ‘Used’ filter on the left sidebar, it worked for others.” The input will come from two branches simultaneously at each call: 

\begin{enumerate}
    \item The Condenser summary of the \emph{current} partial trace (\texttt{summary\_text}${}_t$).
    \item The top-$K$ (\(K=5\)) EMS summaries \( \{\texttt{summary\_text}_{i}\}_{i=1}^{K} \).
\end{enumerate}

With 600 trajectories stored in the EMS, timings showed the cosine loop costs ~10 ms per query regardless of k, so latency doesn’t concern how we choose k’s default value, as shown in Fig~\ref{fig:retrieval_speed}. We pick k=5 because presenting five closest past experiences gives the coach enough varied examples to detect patterns without overwhelming the LLM context window or drowning out the current state.

\paragraph{Decision rule.}

Intervention is selective. The Coach is explicitly trained or prompted to remain silent unless it detects high likelihood of failure (e.g., encountering CAPTCHAs, loops, dead ends) or recognizes a better workflow from memory. This ensures that interventions are timely and meaningful, rather than overwhelming the actor with irrelevant advice. The guidance itself is concise, usually a sentence or two. It is injected into the actor’s prompt as a system message. This injection is entirely non-invasive: no gradients are backpropagated, and the actor’s internal policy remains untouched.

\paragraph{Injection mechanism.}
The LLM is instructed to \emph{intervene} when it predicts either a high failure probability (looping, CAPTCHA, HTTP 4xx), or a faster path exists in memory.  
Otherwise it returns \texttt{"intervene": false}. For \texttt{intervene=true}, the advice JSON is synchronously appended to the actor’s message history as a system message before the next action selection.  
This requires no change to the actor’s policy network.

The Coach offers a scalable way to improve agent reliability without any modifications to the actor’s architecture. It enables continual improvement over time, as more episodes are accumulated and more patterns are learned—all without retraining the original navigation model. By separating summarization (WebCondenser), storage/retrieval (EMS), and decision-making (Coach), WebCoach remains \emph{model- and framework-agnostic}.  Each module is replaceable, enabling rapid experimentation with larger memories, improved retrieval, or custom intervention policies without disturbing the underlying navigation agent.

%% file: sections/3_experiments.tex
\section{Experiments}

\definecolor{darkyellow}{RGB}{204,153,0}
\definecolor{matchagreen}{RGB}{112,148,93}

\begin{figure}[t]
    \centering
    \includegraphics[width=\linewidth]{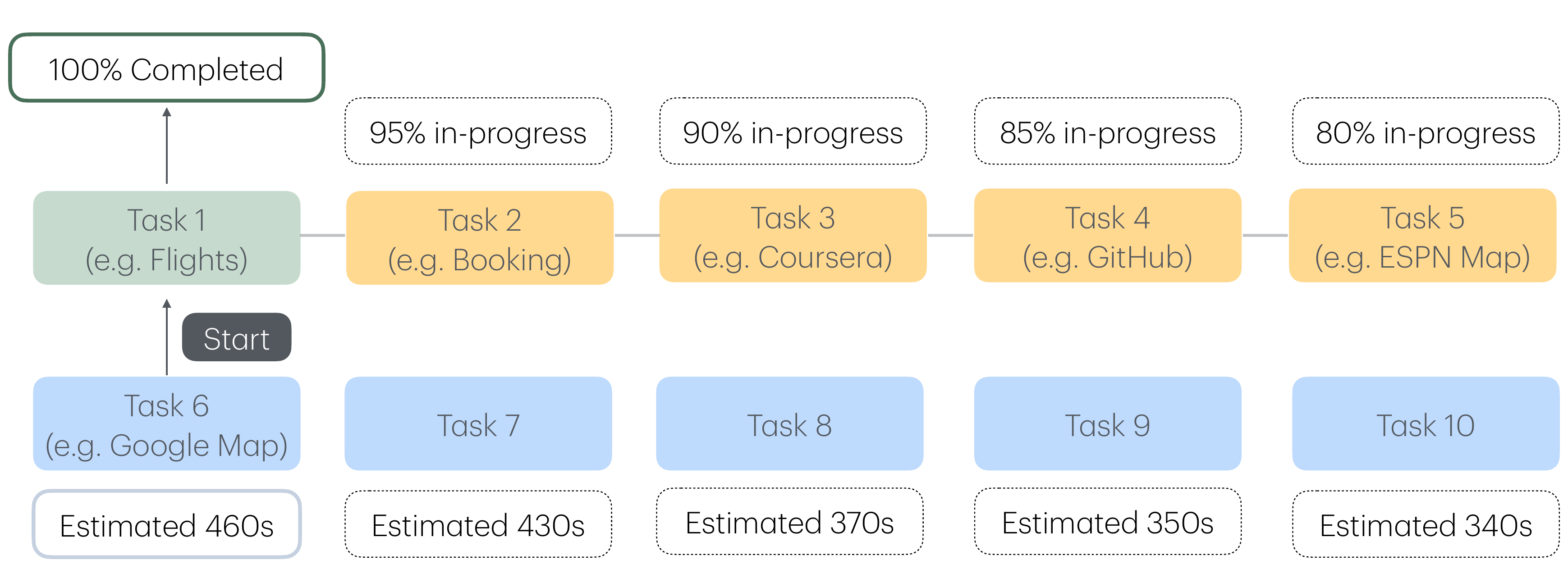}
    \vspace{-4pt}
    \caption{
    \textbf{Asynchronous Evaluation of WebVoyager.}
    WebCoach's asynchronous evaluation pipeline distributes the 15 subdomains in WebVoyager (e.g., \textit{Amazon}, \textit{Apple}, \textit{ArXiv}) across parallel evaluation queues to maximize throughput and GPU utilization. \textcolor{darkyellow}{Yellow} boxes indicates in-progress tasks, \textcolor{matchagreen}{green} indicates completed tasks, and \textcolor{blue}{blue} indicates tasks that are waiting in the queue. Our limited compute supports running 5 tasks in parallel, and once a task finishes earlier than the others in a batch, we immediately start another task from the wait list instead of waiting for the entire batch to finish. This asynchronous queueing strategy reduces total evaluation time by over 80\%, enabling scalable benchmarking of web agents at large scale.
    }
    \label{fig:async_eval}
\end{figure}

\subsection{Data}

We conduct all evaluations in a real browser environment, using \textbf{WebVoyager} \cite{he2024webvoyager}. In contrast to prior work that relies on cached website snapshots \citep{deng2023mind2web, LuKR24} or sandboxed environments with a limited set of simulated sites \citep{yao2022webshop, zhou2023webarena, koh2024visualwebarena}, this benchmark involves live interaction with real-world webpages. This setup reduces the simulation-to-reality gap and better reflects the challenges faced by real users, such as dynamic content, login gates, and UI drift. WebVoyager complements this by enabling large-scale crawling and evaluation across a wide web surface, and remains one of the most widely used platforms for benchmarking web agents under online conditions \citep{he2024webvoyager}.  

By adopting an online benchmark, we ensure that our evaluation aligns with the deployment setting of web agents, where robustness to interface changes, session-level memory, and adaptation to unfamiliar page layouts are crucial. For evaluation, we utilize browser-use agent’s evaluation capability by checking the resulting state of their last action against the initial user query, and determine the success of the task.

\subsection{Setup}

All experiments are conducted in a real Chromium browser environment running inside Docker 
containers. A dedicated image is built for each run, and all dependencies are initialized at container startup to ensure consistent execution across checkpoints and models. Evaluations are performed on an NVIDIA A100 machine. To bound runtime and avoid degenerate 
navigation loops, we enforce a 30-second timeout per action step and a hard cap of 50 steps per task.

\paragraph{Asynchronous Evaluation with Dynamic Batching}

To increase throughput over WebVoyager's 15 subdomains, we adopt a two-level parallelization 
strategy. First, Docker containerization provides isolated browser instances. Second, a Python 
subprocess scheduler groups subdomains into batches and executes several in parallel, while tasks 
within each subdomain are processed sequentially to preserve browser determinism. Webvoyager has 15 distinct subdomains (“Amazon”, “Apple.com”, “Google Flights”, etc) and each subdomain contains 30-50 tasks. In order to maximize evaluation speed and GPU utilization, we serve the open-source VLMs on vLLM and SGLang, and deploy the browser-use agent on chromium browser environment in groups of 5 subdomains at once. As illustrated in Fig~\ref{fig:async_eval}, we maintain an evaluation queue of 5 subdomains, and whenever all tasks from one domain finished running, a new domain will enter the queue and be evaluated, until we process all tasks.

We treat each WebVoyager task as a job with an estimated runtime and schedule them on up to
5 parallel browser workers using an LPT-style (Longest Processing Time first) heuristic: longer tasks are launched earlier, and whenever a worker becomes free it pulls the longest remaining job from a global evaluation queue. This reduces tail latency and shortens the overall makespan compared to naive FIFO batching.

Take Qwen-VL-32b for example. Every task on average took 460s to complete, meaning that for all 643 tasks, a sequential evaluation would take 460*643/3600 = 82 hours to complete. With our parallelization queue strategy the actual running time for this evaluation turned out to be less than 14 hours, yielding a 83\% reduction. 

\paragraph{Base-Agent Configuration.}
The base web agent is a vision-language model (Qwen2.5-VL-7B, Qwen2.5-VL-32B, or Skywork-r1v3-38B) 
served on vLLM or SGLang. Optional chain-of-thought or verbose ``thinking'' modes are disabled unless 
explicitly required by an ablation, as they increase latency without observable gains in navigation 
performance. Apart from WebCoach's injected system messages, we use the default 
\texttt{browseruse} system prompt without modification.

\paragraph{Prompt Usage.}
While the functional roles of the \texttt{WebCondenser} and \texttt{Coach} are described in Section~\ref{sec:webcoach}, here we note only the prompting behavior relevant to evaluation. The Condenser is used strictly for summarizing 
partial and complete trajectories under a fixed schema, and the Coach produces a lightweight JSON-formatted intervention decision. All prompts follow a fixed template to ensure deterministic behavior across runs.

\paragraph{Memory Retrieval and Leakage Control.}
Only completed trajectories are written to the External Memory Store (EMS). During evaluation, 
retrieval explicitly excludes any episode whose WebVoyager task ID matches that of the current task, 
preventing leakage of same-task experience. The EMS returns only experiences from distinct tasks or 
from prior runs of different subtasks. Embedding similarity is computed using normalized dot products, 
and HNSW-128 indexing supports scalable nearest-neighbor search as the memory grows in the dynamic 
setting.

\paragraph{Reproducibility.}
Each run stores its configuration, Condenser outputs, Coach outputs, and (optionally) intermediate 
browser states. These logs enable reproducible replays, ablations, and cross-model comparisons.

%% file: sections/4_results.tex
\section{Results}
\label{sec:results}
\vspace{-3pt}
\begin{figure*}[ht]
    \centering
    \includegraphics[width=\textwidth]{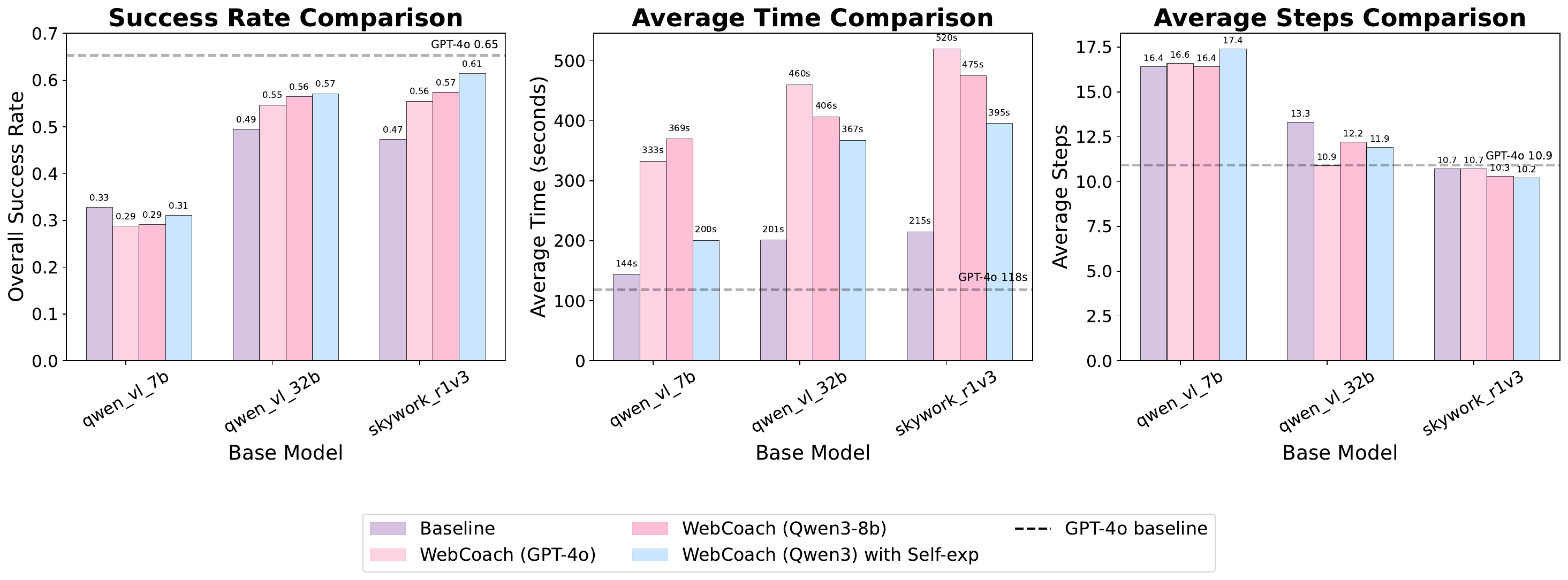}
    \vspace{-6pt}
    \caption{
    \textbf{Performance comparison across base models.}
    WebCoach consistently improves browser-use agents’ reasoning and robustness across different backbones.
    The framework achieves higher success rates (with the exception for the 7b backbone model) while maintaining a roughly equal or fewer average steps.
    }
    \label{fig:webcoach_results}
\end{figure*}

\begin{table}[ht]
\centering
\small
\resizebox{\linewidth}{!}{
\begin{tabular}{l l l c c c}
\toprule
\textbf{Memory} & \textbf{Coach Model} & \textbf{Base Model} 
& \textbf{Avg. Time (s)} & \textbf{Avg. Steps} & \textbf{Success Rate} \\
\midrule
\rowcolor{gray!10}

None & None & GPT-4o & 118 & 10.9 & 0.653 \\
None & None & Qwen-VL-7B & 144 & 16.4 & 0.328 \\
None & None & Qwen-VL-32B & 201 & 13.3 & 0.495 \\
None & None & Skywork-38B & 215 & 10.7 & 0.473 \\
\midrule
External (Frozen) & GPT-4o & Qwen-VL-7B & 333 & 16.6 & 0.288 \\
External (Frozen) & GPT-4o & Qwen-VL-32B & 460 & 10.9 & 0.547 \\
External (Frozen) & GPT-4o & Skywork-38B & 520 & 10.7 & 0.555 \\
\midrule
External (Frozen) & Qwen3-8B & Qwen-VL-7B & 369 & 16.4 & 0.291 \\
External (Frozen) & Qwen3-8B & Qwen-VL-32B & 406 & 12.2 & 0.565 \\
External (Frozen) & Qwen3-8B & Skywork-38B & 475 & 10.3 & 0.574 \\
\midrule
Self (Dynamic) & Qwen3-8B & Qwen-VL-7B & 200 & 17.4 & 0.311 \\
Self (Dynamic) & Qwen3-8B & Qwen-VL-32B & 367 & 11.9 & 0.571 \\
Self (Dynamic) & Qwen3-8B & Skywork-38B & 395 & 10.2 & \textbf{0.614} \\
\bottomrule
\end{tabular}}
\vspace{-3pt}
\caption{Success rate, average time, and average number of steps to completion across WebVoyager benchmark experiments. Columns reorganized to show memory type and coach model explicitly.}
\label{tab:webcoach_results}
\end{table}

We conducted four sets of experiments on WebVoyager \citep{he2024webvoyager} across three base models: \texttt{Qwen2.5-VL-7B}, \texttt{Qwen2.5-VL-32B}, and \texttt{Skywork-r1v3-38B}, alongside a ceiling baseline with \texttt{GPT-4o}. Each run covered the full WebVoyager benchmark of 643 online tasks. The experiment configurations were:

\begin{enumerate}
    \item Baseline: no coaching enabled.
    \item Frozen EMS (GPT-4o coach): memory initialized with GPT-4o trajectories and GPT-4o acting as the coach.
    \item Frozen EMS (Qwen3-8B coach): memory initialized with GPT-4o trajectories but coached by Qwen3-8B.
    \item Dynamic EMS (Qwen3-8B coach): each main agent iteratively updates its own trajectories into the memory store, coached by Qwen3-8B.
\end{enumerate}

Interestingly, GPT-4o did not consistently outperform Qwen3-8B as a coach, which led us to decide against DPO-based finetuning of Qwen3. We believe that Qwen3-8B is sufficiently capable to act as a condenser and coach in the zero-shot setting, since this task primarily depends on trajectory-level reasoning rather than complex instruction following.

\vspace{4pt}
\noindent
\textbf{Quantitative analysis.}
Figure~\ref{fig:webcoach_results} and Table~\ref{tab:webcoach_results} reveal a consistent pattern of improvement across both success rate and efficiency metrics. The largest jump occurs in the Skywork-38B model, which improves from 47.3\% to 61.4\% success rate when equipped with WebCoach, a 14.4-point gain, comparable to the GPT-4o ceiling baseline. The Qwen-VL-32B also improves by over 7 points (49.5\% $\rightarrow$ 57.1\%), confirming that experience-guided coaching scales effectively with model capacity. In all cases, the number of action steps per task remains flat or decreases, suggesting that the observed gains stem not from brute-force exploration but from better-informed decision paths. 

While average completion time rises due to the additional inference overhead of the Coach and Condenser, the efficiency gain in navigation behavior outweighs the latency cost. For instance, dynamic EMS with self-expansion adds $\sim$150 seconds of average runtime relative to baseline, yet reduces redundant actions by 1–2 steps per episode. The overhead is also transient: as the EMS grows, HNSW's logarithmic search complexity ensures that retrieval remains highly efficient.

\vspace{4pt}
\noindent
\textbf{Effect of memory source and self-experience.}
We observe that self-generated experiences yield more transferable and actionable knowledge than foreign demonstrations. When agents iteratively grow their own EMS, they encode decision patterns consistent with their own inductive biases and representation space. For example, a Qwen-32B agent retrieves embeddings closer to its own prior successful action embeddings, resulting in smoother reasoning continuity across tasks. In contrast, GPT-4o-derived trajectories, while high-quality, occasionally inject stylistic mismatches. This explains why the dynamic EMS variant achieves higher success with lower step counts than frozen EMS, despite starting from an empty database.

\vspace{4pt}
\noindent
\textbf{Scaling effects and cognitive thresholds.}
The benefits of WebCoach are strongly correlated with model scale. Both Skywork-38B and Qwen-VL-32B show pronounced gains, while the 7B backbone does not benefit, with its success rate slightly decreasing (0.328-0.311). We hypothesize that there is a \emph{cognitive threshold}, where smaller models lack the grounding and reasoning depth to exploit cross-episode memory effectively. Once models approach the reasoning frontier—able to nearly solve a task on their own—the coaching signal serves as a crucial disambiguator that helps avoid high-level decision errors (e.g., repeated scrolling, redundant navigation loops). Thus, memory guidance becomes more valuable at the boundary of partial competence rather than total ignorance.

\vspace{4pt}
\noindent
\textbf{Behavioral analysis.}
Qualitative inspection of logged traces further supports these trends. Agents equipped with WebCoach exhibit clearer high-level planning and fewer repetitive page visits. For instance, on the \textit{Apple.com} and \textit{Amazon} subtasks, coached agents learned to skip redundant login or trade-in prompts that previously caused deadlocks. In dynamic EMS mode, the coach often cited earlier self-generated traces when warning about such pitfalls—demonstrating retrieval-based reflection. In contrast, baseline agents frequently cycled between similar UI elements without learning from past errors. We have more visual recap of what our pipeline does at different steps of a web navigation task in Appendix~\ref{app:visual-example}.

At the subdomain level, the largest improvements appear in semantically complex sites like \textit{Apple}, \textit{ArXiv}, and \textit{BBC News}, where reasoning and element disambiguation are key. Simpler domains such as \textit{Booking.com} and \textit{Google Flights} show smaller or negligible gains, suggesting that memory is most beneficial for multi-step information extraction tasks rather than atomic button-clicking ones.

\vspace{4pt}
\noindent
\textbf{Summary.}
Overall, the results highlight that (1) WebCoach enhances task robustness through evidence-grounded reflection, (2) self-evolving EMS accelerates adaptation without external demonstrations, and (3) scaling the base agent amplifies the value of memory-guided reasoning. Together, these findings demonstrate that persistent, cross-session memory can close much of the performance gap between open-source and proprietary LLM agents on complex, dynamic web environments.

%% file: sections/5_related_work.tex
\section{Related Work}
\label{sec:related_work}

\paragraph{Reasoning-Centric Web and GUI Agents}

Research on interactive agents has advanced along three converging directions.
First, GUI-centric studies combine reward shaping, curriculum design, and self-reflection to turn small or open-source VLMs into capable mobile or desktop controllers \citep{lu2025ui, lian2025ui, shi2025mobilegui, bai2024digirl, liu2025infigui}.
These studies often leverage automated supervision through goal synthesis, self-critique, and cross-platform action libraries to reduce annotation cost and boost generalization \citep{yang2025zerogui, wu2025gui, xu2024aguvis, wu2024atlas}.
Second, work on web navigation integrates multi-turn RL, structured exploration, and hierarchical planning, improving performance on benchmarks like WebArena and WebShop while underscoring the importance of strong visual grounding \citep{wei2025webagent, zhang2025shop, gandhi2025go, li2025websailor, yang2024agentoccam, putta2024agent, zheng2024gpt, nakano2021webgpt, gur2023real, lee2025learning}.
Finally, platform-level systems pursue generality by unifying planning, tool use, and self-evolution to automate complex desktop or Windows workflows with minimal task-specific engineering \citep{agashe2024agent, zhang2024ufo, qiu2025alita}.

\paragraph{Agentic Memory and Context Management}

Work on memory has evolved from simple history compression to scalable, structured substrates. While even simple summarization of past states improves automation accuracy, more advanced multimodal agents employ dual episodic-semantic stores that reduce storage while increasing precision \citep{zhu2025turbocharging, zhang2025ella, wang2025mirix}. Other approaches draw on operating-system analogies to organize information into short-, mid-, and long-term layers or action logs, enhancing performance across language, code, and GUI tasks \citep{kang2025memory, tang2025agent, gao2025chain}.~Techniques like latent-space retrieval and Zettelkasten-style linking further extend reasoning across hundreds of thousands of tokens \citep{wang2025m+, yu2025memagent, xu2025mem}. Workflow-oriented stores mine reusable action sequences to improve web and multi-agent performance \citep{wang2024agent, rahman2025x, liu2025mosaic}, while production systems achieve constant-memory operation with lower latency and strong downstream gains \citep{zhou2025mem1, chhikara2025mem0}. These directions together indicate a shift toward adaptive, long-horizon memory for multimodal and collaborative agents.

\paragraph{Self-Evolving Agents}

Self-evolving agents refine themselves through experience rather than retraining.
Foundational techniques include replay buffers, natural-language reflection, and automatic curricula, which improve performance in both web and embodied environments \citep{liu2025contextual, shinn2023reflexion, wang2023voyager, ouyang2025reasoningbank}.
Subsequent efforts discover reusable skills or APIs via program synthesis for cross-task transfer \citep{wang2025inducing, zheng2025skillweaver}, while self-generated trajectories and scalable curricula close the gap between open-source and proprietary models \citep{qi2024webrl, patel2024large}.
Recent extensions integrate world-model imagination and transition abstraction for long-horizon or mobile settings \citep{fang2025webevolver, chae2024web, wang2025mobile}.
A comprehensive survey summarizes these evolving strategies and highlights open challenges for continual self-improvement \citep{gao2025survey}.

%% file: sections/6_conclusion.tex
\section{Conclusion}
We presented WebCoach, a model-agnostic, self-evolving framework that augments web browsing agents with persistent, cross-session memory. By integrating a lightweight WebCondenser, a scalable External Memory Store (EMS), and a retrieval-based Coach, WebCoach enables existing agents to reflect on prior trajectories, generalize from experience, and recover from repeated errors without retraining. Through large-scale online evaluations on WebVoyager, WebCoach demonstrated consistent gains across three open-source LLM backbones, while maintaining comparable or lower average step counts. These results underscore the value of experience-driven memory in improving robustness and efficiency in web navigation. Future work will (1) integrate the external memory directly into the agent's internal policy to eliminate multi-LLM dependency, and (2) explore reinforcement-based optimization over long-term reward signals. More broadly, our findings highlight that memory-centric design is an essential step towards building continually improving, self-reflective web agents.


%% file: sections/7_appendix.tex
\section{Appendix A: Step-by-Step Visual Illustrion of WebCoach’s Operation}
\label{app:visual-example}

\begin{takeaway}
    
\textcolor{takeawaycolor}{\textbf{Task Description:}} 
On Apple's website, check if the HomePod mini in store is available in multiple colors and list them. Website: https://www.apple.com/

\paragraph{Coaching} Coaching output is a json file including whether or not to intervene, and if so, the exact actionable advice message. For debugging purposes, we also present the retrieved experiences that include meta information, a summary, and current patterns with the specific web elements that the agent interacted with. At this step the coach detected some challenges and decides to issue an advice:

\textcolor{takeawaycolor}{\textbf{[Coaching advice output]}} 

Based on past experiences, it's important to avoid getting stuck in loops or repeatedly failing to apply specific filters. Instead of navigating through the site without a clear path, look for a 'Choose your color' section or dropdown menu on the HomePod mini product page. If you can't find it, consider using the search function with specific terms like 'HomePod mini color options' or 'HomePod mini available colors' to directly locate the information. If you encounter any navigation issues or loops, immediately return to the product page and check for color options there.

\textcolor{takeawaycolor}{\textbf{[Retrieved Relevant Experience 1]}} 

Summary: The user is attempting to find information about the MacBook Air color options on Apple's website. They started by scrolling the homepage, clicked on a MacBook Pro link, and then navigated through several unrelated pages like AirPods and iPhone. After using the search bar to find 'MacBook Air', they reached the product page but have been scrolling and searching without success in locating the color options section. The user has attempted various methods, including using the search function and looking for a 'Customize' button, but has not yet found the relevant information.

\textcolor{takeawaycolor}{\textbf{[Current Patterns]}} 

\begin{enumerate}

    \item Repeated Navigation to MacBook Air Page: The user repeatedly navigates to the MacBook Air product page, likely due to a misstep or a need to verify the correct page.

    \item Persistent Scrolling on MacBook Air Page: The user continues to scroll down the MacBook Air product page in an effort to find the color options section, indicating that they haven't found the information yet.

    \item Repeated Search Bar Interactions: The user frequently interacts with the search bar, possibly due to confusion about the page structure or uncertainty about the location of the color options.

\end{enumerate}

Similarity score: 0.8895191550254822

[2 more relevant retrieved experiences omitted]

\textcolor{takeawaycolor}{\textbf{[Main agent returned results:]}} 

is\_done: true,

success: true,

long-term memory: Task completed: True - The available colors for the HomePod mini are: Blue, Yellow, Midnight, White, and Orange.

\textcolor{takeawaycolor}{\textbf{[Condenser output (to be stored in the EMS)]}} 

Summary: The user successfully completed the task of checking the availability of the HomePod mini in multiple colors. They navigated from the Apple homepage to the TV and Home section, then to the HomePod mini product page, and finally identified the color options. Although there were some steps where the user scrolled and searched for color options, they eventually found the color selection on the product page.

\textbf{Success workflows}:

\begin{enumerate}
    \item name: Navigation to Product Page,
      description: The user navigated from the Apple homepage to the TV and Home section and then to the HomePod mini product page, which is a key step in accessing the product information.

    \item name: Color Option Identification,
      description: The user successfully identified the available colors (Blue, Yellow, Midnight, White, and Orange) on the product page, which is the main goal of the task.

    \item name: Scrolling and Searching for Color Options,
      description: The user attempted to locate the color options by scrolling through the product page, which eventually led them to the color selection section.
\end{enumerate}

embedding: (...omitted...)

\end{takeaway}

\begin{figure*}[!htbp]
    \centering
    \includegraphics[width=0.8\textwidth]{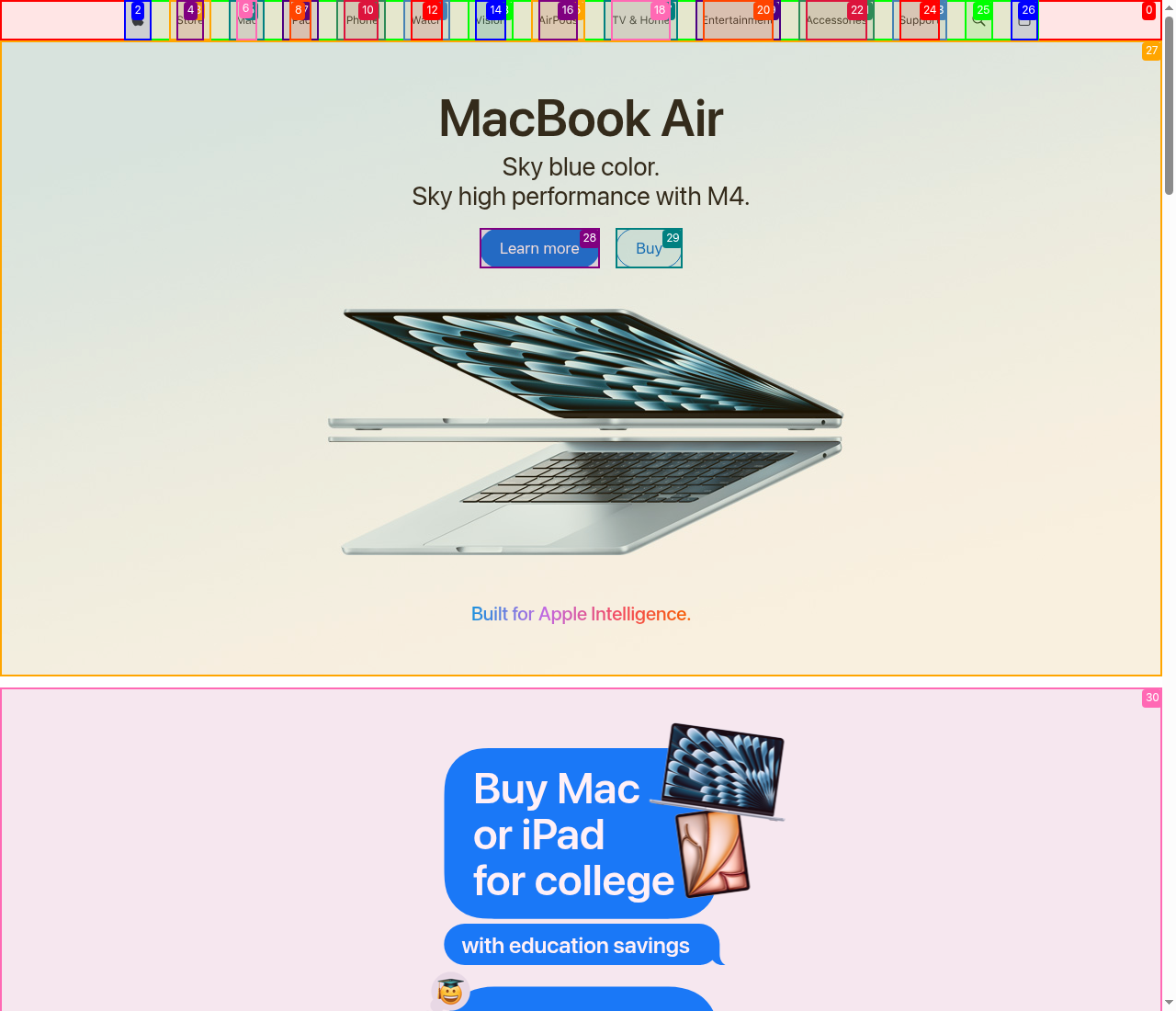}
    \caption{Step 1 Screenshot}
\end{figure*}

\begin{figure*}[!htbp]
    \centering
    \includegraphics[width=0.8\textwidth]{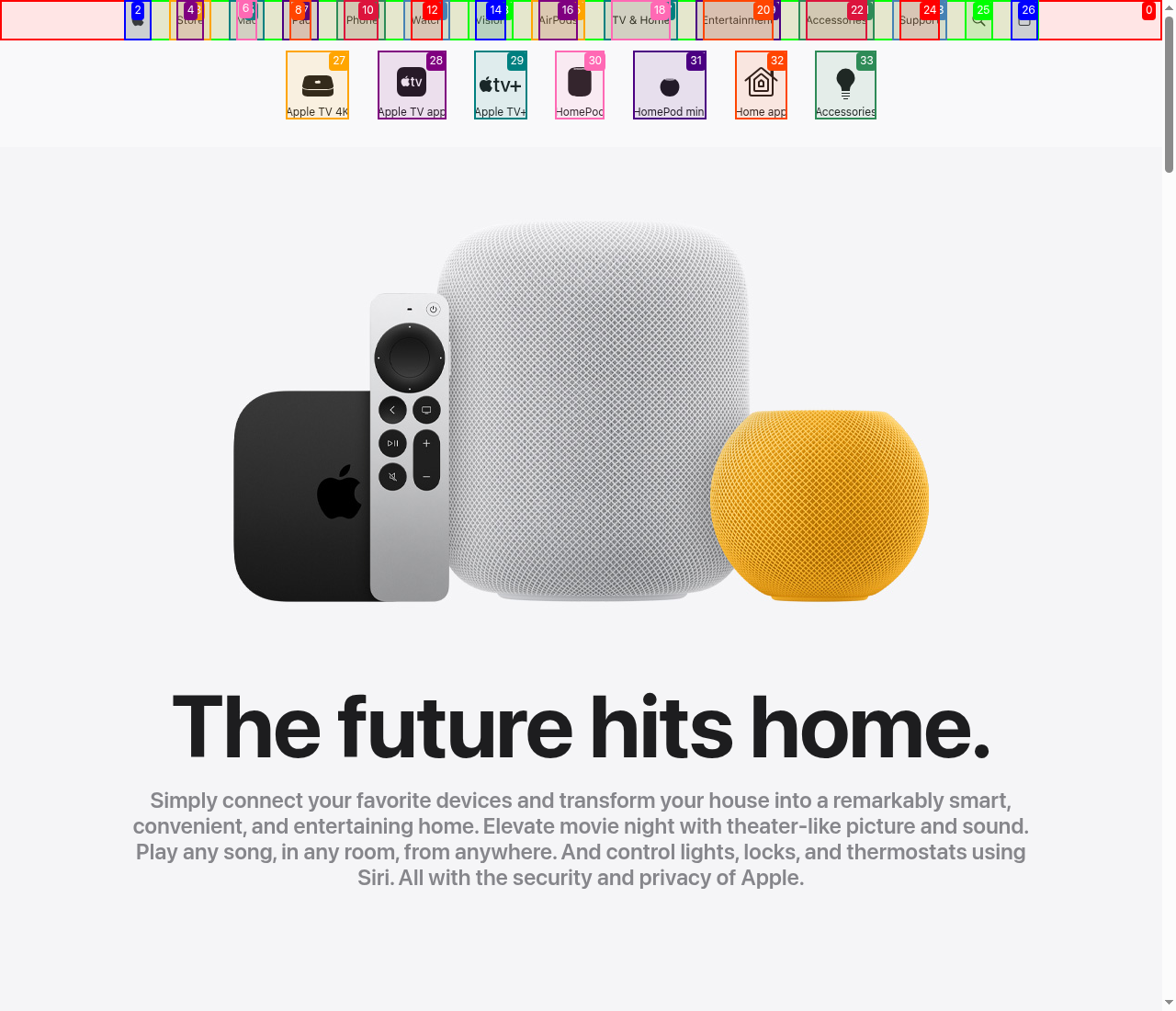}
    \caption{Step 2 Screenshot}
\end{figure*}

\begin{figure*}[!htbp]
    \centering
    \includegraphics[width=0.8\textwidth]{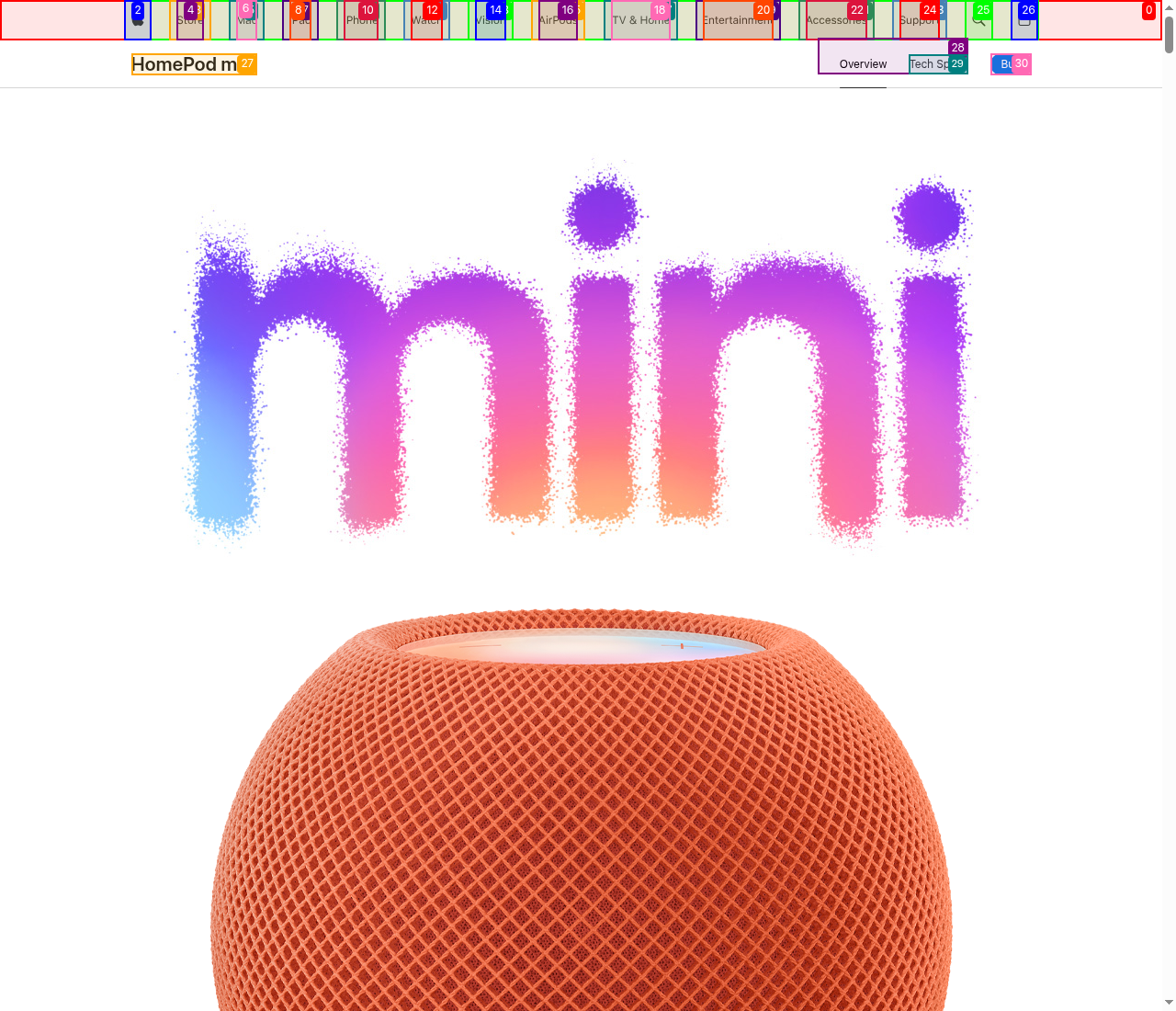}
    \caption{Step 3 Screenshot}
\end{figure*}

\begin{figure*}[!htbp]
    \centering
    \includegraphics[width=0.8\textwidth]{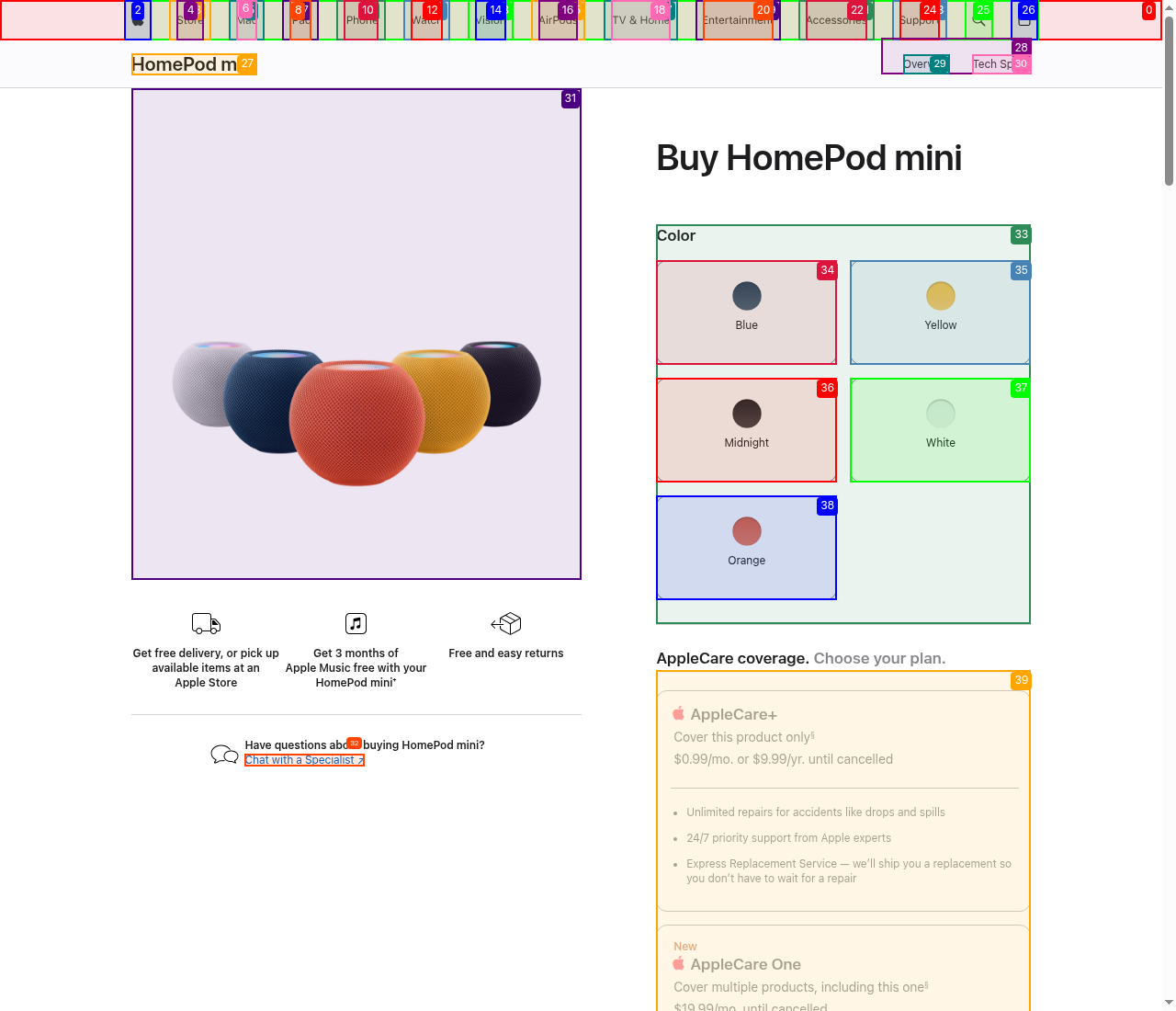}
    \caption{Step 4 Screenshot}
\end{figure*}

\FloatBarrier
\section{Appendix B: Subtask Results}

\input{tables/baseline-gpt4o}

\input{tables/with_coach_4o_exp_4o_coach}

\input{tables/with_coach_4o_exp_qwen3_coach}

\input{tables/with_coach_self_evolve_qwen3_coach}

%% file: tables/baseline-gpt4o.tex
\begin{table*}[!htbp]
\centering
\scriptsize
\setlength{\tabcolsep}{2pt}
\begin{tabular}{lcccccccccccccccc}
\toprule
\multirow{2}{*}{\textbf{Subtask}} &
\multicolumn{4}{c}{\textbf{GPT-4o}} &
\multicolumn{4}{c}{\textbf{Qwen-VL-32B}} &
\multicolumn{4}{c}{\textbf{Qwen-VL-7B}} &
\multicolumn{4}{c}{\textbf{Skywork-r1v3}} \\
\cmidrule(lr){2-5}\cmidrule(lr){6-9}\cmidrule(lr){10-13}\cmidrule(lr){14-17}
& SR & S/T & Time & Steps & SR & S/T & Time & Steps & SR & S/T & Time & Steps & SR & S/T & Time & Steps \\
\midrule
Allrecipes           & 0.889 & 40/45 & 67  & 7.4  & 0.644 & 29/45 & 254 & 12.3 & 0.356 & 16/45 & 142 & 16.1 & 0.556 & 25/45 & 204 & 11.0 \\
Amazon               & 0.951 & 39/41 & 92  & 6.5  & 0.707 & 29/41 & 210 & 11.1 & 0.659 & 27/41 & 193 & 17.1 & 0.463 & 19/41 & 195 & 9.1 \\
Apple                & 0.605 & 26/43 & 119 & 11.7 & 0.558 & 24/43 & 264 & 12.9 & 0.209 & 9/43  & 187 & 15.6 & 0.488 & 21/43 & 209 & 10.5 \\
ArXiv                & 0.837 & 36/43 & 77  & 8.0  & 0.628 & 27/43 & 153 & 10.8 & 0.233 & 10/43 & 166 & 15.0 & 0.698 & 30/43 & 150 & 7.2 \\
BBC News             & 0.714 & 30/42 & 87  & 10.3 & 0.571 & 24/42 & 159 & 12.0 & 0.190 & 8/42  & 141 & 14.3 & 0.690 & 29/42 & 181 & 9.4 \\
Booking              & 0.136 & 6/44  & 303 & 18.4 & 0.068 & 3/44  & 398 & 18.6 & 0.182 & 8/44  & 162 & 17.5 & 0.045 & 2/44  & 297 & 12.7 \\
Cambridge Dict.      & 0.953 & 41/43 & 48  & 5.3  & 0.837 & 36/43 & 92  & 7.6  & 0.628 & 27/43 & 115 & 13.9 & 0.721 & 31/43 & 119 & 6.4 \\
Coursera             & 0.595 & 25/42 & 109 & 12.9 & 0.714 & 30/42 & 177 & 13.3 & 0.452 & 19/42 & 149 & 16.3 & 0.571 & 24/42 & 254 & 13.5 \\
ESPN                 & 0.659 & 29/44 & 114 & 11.7 & 0.432 & 19/44 & 205 & 14.2 & 0.182 & 8/44  & 178 & 17.9 & 0.455 & 20/44 & 201 & 9.5 \\
GitHub               & 0.854 & 35/41 & 82  & 8.5  & 0.537 & 22/41 & 157 & 13.8 & 0.317 & 13/41 & 118 & 16.9 & 0.610 & 25/41 & 274 & 12.9 \\
Google Flights       & 0.071 & 3/42  & 200 & 19.3 & 0.095 & 4/42  & 249 & 18.6 & 0.310 & 13/42 & 169 & 18.9 & 0.071 & 3/42  & 370 & 16.5 \\
Google Map           & 0.878 & 36/41 & 81  & 8.2  & 0.561 & 23/41 & 141 & 11.6 & 0.463 & 19/41 & 112 & 15.8 & 0.610 & 25/41 & 205 & 10.4 \\
Google Search        & 0.233 & 10/43 & 192 & 15.6 & 0.070 & 3/43  & 249 & 17.5 & 0.279 & 12/43 & 114 & 19.0 & 0.279 & 12/43 & 181 & 9.9 \\
Huggingface          & 0.674 & 29/43 & 123 & 11.3 & 0.512 & 22/43 & 169 & 13.9 & 0.279 & 12/43 & 107 & 15.0 & 0.581 & 25/43 & 214 & 11.1 \\
Wolfram Alpha        & 0.761 & 35/46 & 79  & 8.3  & 0.522 & 24/46 & 130 & 11.9 & 0.435 & 20/46 & 108 & 17.0 & 0.283 & 13/46 & 178 & 9.9 \\
\midrule
\textbf{Overall}     & \textbf{0.653} & \textbf{420/643} & \textbf{118.4} & \textbf{10.9}
                     & \textbf{0.495} & \textbf{319/643} & \textbf{200.9} & \textbf{13.3}
                     & \textbf{0.344} & \textbf{221/643} & \textbf{144.0} & \textbf{16.4}
                     & \textbf{0.473} & \textbf{304/643} & \textbf{215.0} & \textbf{10.7} \\
\bottomrule
\end{tabular}
\caption{Baseline: Without WebCoach, per-subtask comparison across base models on the \textbf{WebVoyager} benchmark. 
Each model’s subcolumns report Success Rate (SR), number of successful vs.\ total tasks (S/T), average completion time (s), and average action steps. GPT-4o serves as the ceiling reference.}
\label{tab:per_subtask_comparison}
\end{table*}

%% file: tables/with_coach_4o_exp_4o_coach.tex
\begin{table*}[!htbp]
\centering
\scriptsize
\setlength{\tabcolsep}{5pt}
\begin{tabular}{lcccccccccccccccc}
\toprule
\multirow{2}{*}{\textbf{Subtask}} &
\multicolumn{4}{c}{\textbf{Qwen-VL-32B}} &
\multicolumn{4}{c}{\textbf{Qwen-VL-7B}} &
\multicolumn{4}{c}{\textbf{Skywork-r1v3}} \\
\cmidrule(lr){2-5}\cmidrule(lr){6-9}\cmidrule(lr){10-13}
& SR & S/T & Time & Steps & SR & S/T & Time & Steps & SR & S/T & Time & Steps \\
\midrule
Allrecipes           & 0.644 & 29/45 & 337 & 12.2 & 0.311 & 14/45 & 292 & 16.5 & 0.600 & 27/45 & 561 & 12.5 \\
Amazon               & 0.805 & 33/41 & 370 & 11.8 & 0.585 & 24/41 & 370 & 17.4 & 0.829 & 34/41 & 556 & 11.1 \\
Apple                & 0.395 & 17/43 & 353 & 13.0 & 0.233 & 10/43 & 347 & 14.8 & 0.395 & 17/43 & 667 & 13.7 \\
ArXiv                & 0.698 & 30/43 & 282 & 9.9  & 0.209 & 9/43  & 312 & 14.4 & 0.721 & 31/43 & 322 & 7.1  \\
BBC News             & 0.595 & 25/42 & 311 & 12.8 & 0.143 & 6/42  & 346 & 16.7 & 0.667 & 28/42 & 479 & 10.4 \\
Booking              & 0.045 & 2/44  & 910 & 15.5 & 0.182 & 8/44  & 378 & 17.7 & 0.091 & 4/44  & 858 & 14.8 \\
Cambridge Dict.      & 0.907 & 39/43 & 261 & 6.1  & 0.605 & 26/43 & 261 & 13.9 & 0.953 & 41/43 & 273 & 6.1  \\
Coursera             & 0.690 & 29/42 & 637 & 13.2 & 0.357 & 15/42 & 348 & 17.7 & 0.738 & 31/42 & 535 & 11.3 \\
ESPN                 & 0.591 & 26/44 & 467 & 8.8  & 0.114 & 5/44  & 382 & 17.6 & 0.614 & 27/44 & 460 & 9.1  \\
GitHub               & 0.829 & 34/41 & 520 & 10.4 & 0.195 & 8/41  & 294 & 17.0 & 0.829 & 34/41 & 526 & 10.6 \\
Google Flights       & 0.071 & 3/42  & 926 & 17.5 & 0.119 & 5/42  & 367 & 18.0 & 0.095 & 4/42  & 882 & 17.8 \\
Google Map           & 0.732 & 30/41 & 461 & 9.7  & 0.463 & 19/41 & 309 & 16.1 & 0.707 & 29/41 & 426 & 9.5  \\
Google Search        & 0.233 & 10/43 & 314 & 7.0  & 0.140 & 6/43  & 336 & 18.3 & 0.302 & 13/43 & 413 & 9.0  \\
Huggingface          & 0.674 & 29/43 & 425 & 8.9  & 0.163 & 7/43  & 317 & 15.3 & 0.605 & 26/43 & 460 & 10.1 \\
Wolfram Alpha        & 0.348 & 16/46 & 341 & 7.5  & 0.500 & 23/46 & 330 & 17.0 & 0.239 & 11/46 & 383 & 8.3  \\
\midrule
\textbf{Overall}     & \textbf{0.547} & \textbf{352/643} & \textbf{460.1} & \textbf{10.9}
                     & \textbf{0.288} & \textbf{185/643} & \textbf{332.6} & \textbf{16.6}
                     & \textbf{0.555} & \textbf{357/643} & \textbf{519.6} & \textbf{10.7} \\
\bottomrule
\end{tabular}
\caption{Per-subtask results \textbf{with WebCoach} (GPT-4o experiences, frozen EMS, GPT-4o as coach). 
Each subtask corresponds to a specific website domain. 
Metrics include success rate (SR), successful vs.\ total tasks (S/T), average completion time (s), and average action steps.}
\label{tab:webcoach_frozen_gpt4o}
\end{table*}

%% file: tables/with_coach_4o_exp_qwen3_coach.tex
\begin{table*}[!htbp]
\centering
\scriptsize
\setlength{\tabcolsep}{5pt}
\begin{tabular}{lcccccccccccccccc}
\toprule
\multirow{2}{*}{\textbf{Subtask}} &
\multicolumn{4}{c}{\textbf{Qwen-VL-32B}} &
\multicolumn{4}{c}{\textbf{Qwen-VL-7B}} &
\multicolumn{4}{c}{\textbf{Skywork-r1v3}} \\
\cmidrule(lr){2-5}\cmidrule(lr){6-9}\cmidrule(lr){10-13}
& SR & S/T & Time & Steps & SR & S/T & Time & Steps & SR & S/T & Time & Steps \\
\midrule
Allrecipes           & 0.644 & 29/45 & 305 & 11.7 & 0.267 & 12/45 & 328 & 16.3 & 0.711 & 32/45 & 406 & 10.0 \\
Amazon               & 0.805 & 33/41 & 276 & 9.9  & 0.488 & 20/41 & 349 & 15.0 & 0.732 & 30/41 & 477 & 9.7  \\
Apple                & 0.488 & 21/43 & 402 & 14.5 & 0.233 & 10/43 & 354 & 15.8 & 0.605 & 26/43 & 460 & 10.5 \\
ArXiv                & 0.721 & 31/43 & 261 & 10.0 & 0.233 & 10/43 & 290 & 14.3 & 0.791 & 34/43 & 268 & 5.9  \\
BBC News             & 0.571 & 24/42 & 259 & 12.5 & 0.190 & 8/42  & 320 & 15.5 & 0.619 & 26/42 & 411 & 9.5  \\
Booking              & 0.045 & 2/44  & 869 & 17.2 & 0.182 & 8/44  & 485 & 18.1 & 0.068 & 3/44  & 870 & 15.6 \\
Cambridge Dict.      & 0.930 & 40/43 & 253 & 6.0  & 0.558 & 24/43 & 372 & 13.5 & 0.837 & 36/43 & 241 & 6.1  \\
Coursera             & 0.690 & 29/42 & 553 & 12.1 & 0.333 & 14/42 & 510 & 18.0 & 0.690 & 29/42 & 589 & 12.6 \\
ESPN                 & 0.614 & 27/44 & 507 & 10.5 & 0.159 & 7/44  & 437 & 16.7 & 0.545 & 24/44 & 456 & 9.9  \\
GitHub               & 0.805 & 33/41 & 447 & 9.3  & 0.220 & 9/41  & 445 & 18.4 & 0.878 & 36/41 & 412 & 9.2  \\
Google Flights       & 0.119 & 5/42  & 538 & 18.4 & 0.214 & 9/42  & 377 & 18.2 & 0.095 & 4/42  & 892 & 18.0 \\
Google Map           & 0.732 & 30/41 & 279 & 10.8 & 0.390 & 16/41 & 318 & 16.0 & 0.756 & 31/41 & 412 & 9.3  \\
Google Search        & 0.070 & 3/43  & 504 & 17.3 & 0.116 & 5/43  & 342 & 18.5 & 0.279 & 12/43 & 401 & 9.1  \\
Huggingface          & 0.535 & 23/43 & 359 & 12.6 & 0.233 & 10/43 & 320 & 16.1 & 0.651 & 28/43 & 435 & 10.0 \\
Wolfram Alpha        & 0.717 & 33/46 & 277 & 9.8  & 0.543 & 25/46 & 299 & 15.9 & 0.391 & 18/46 & 400 & 8.7  \\
\midrule
\textbf{Overall}     & \textbf{0.565} & \textbf{363/643} & \textbf{406.4} & \textbf{12.2}
                     & \textbf{0.291} & \textbf{187/643} & \textbf{369.4} & \textbf{16.4}
                     & \textbf{0.574} & \textbf{369/643} & \textbf{475.0} & \textbf{10.3} \\
\bottomrule
\end{tabular}
\caption{Per-subtask results \textbf{with WebCoach} (GPT-4o experiences, frozen EMS, Qwen3-8B as coach). 
Each model’s subcolumns report success rate (SR), successful vs.\ total tasks (S/T), average completion time (s), and average action steps.}
\label{tab:webcoach_frozen_qwen3}
\end{table*}